\documentclass[runningheads]{llncs}
\usepackage[utf8]{inputenc}
\usepackage{amsmath, amssymb}
\usepackage[ruled,vlined,linesnumbered]{algorithm2e}
\usepackage{multirow}
\usepackage{tcolorbox}
\usepackage{makecell}
\usepackage{algpseudocode}
\usepackage{amsmath}
\usepackage[T1]{fontenc}

\usepackage{tikz}
\usepackage{pgfplots}
\pgfplotsset{compat=1.18}
\usepackage{booktabs}
\usepackage{xcolor}
\usepackage{graphicx}
\begin{document}
\title{D$^2$F-ReAG: Dynamic Decomposition and Filtering for Multi-Hop Reasoning-Augmented Generation}
%
%
\author{Jiaoyang Li\inst{1} \and
Junhao Ruan\inst{1} \and
Shengwei Tang\inst{1} \and
Saihan Chen\inst{1} \and
Kaiyan Chang\inst{1} \and
Zhengtao Yu\inst{2} \and
Tong Xiao\inst{1,3} \and
Jingbo Zhu\inst{1,3}}
\authorrunning{J. Li et al.}
%
\author{Jiaoyang Li\inst{1} \and
Junhao Ruan\inst{1} \and
Shengwei Tang\inst{1} \and
Saihan Chen\inst{1} \and
Kaiyan Chang\inst{1} \and
Zhengtao Yu\inst{2} \and
Tong Xiao\inst{1,3} \and
Jingbo Zhu\inst{1,3}}
\authorrunning{J. Li et al.}
\institute{Northeastern University, Shenyang, China \and
Kunming University of Science and Technology, Kunming, China \and
NiuTrans Research, Shenyang, China}
\maketitle              

\begin{abstract}
Large language models (LLMs) often generate inaccurate answers due to their reliance on static internal knowledge. Retrieval-augmented generation (RAG) addresses this limitation by integrating external knowledge and excelling at single-hop queries. However, it struggles with multi-hop questions that require cross-document reasoning. Existing methods, such as graph structured RAG or question decomposition, often lack dynamic decomposition and effective filtering, which leads to lower efficiency and accuracy. 
To overcome these limitations, we propose \textbf{D}ynamic \textbf{D}ecomposition and \textbf{F}iltering for Multi-Hop \textbf{Re}asoning-\textbf{A}ugmented \textbf{G}eneration (D$^2$F-ReAG), a novel paradigm that adaptively controls reasoning depth by judging the reliability of the root-level reasoning. If the root reasoning is reliable, the model directly generates the answer. Otherwise, the question is logically decomposed into sub-questions, and the verified reasoning derived from these sub-questions is used to refine the root reasoning. Experiments on three multi-hop benchmarks demonstrate the effectiveness of our method in handling complex multi-hop questions.


\end{abstract}
\section{Introduction}

Large Language Models (LLMs) have demonstrated strong capabilities in language understanding and generation across a wide range of tasks \cite{achiam2023gpt}. Despite these advances, they still suffer from inherent limitations such as outdated parametric knowledge and hallucinations, which often lead to factually incorrect or unsupported outputs and undermine their reliability in knowledge-intensive scenarios \cite{jiang-etal-2024-large}. Retrieval-augmented Generation (RAG) mitigates these issues by incorporating external knowledge sources into the generation process, thereby improving factuality and reducing the model's reliance on its static internal knowledge \cite{lewis2020retrieval}.

For simple factual questions, retrievers can often locate the necessary evidence within a single retrieval step, making the process both efficient and straightforward \cite{shi-etal-2024-generate}. In contrast, multi-hop reasoning questions require connecting multiple pieces of evidence that are typically scattered across different documents\cite{cheng-etal-2025-dualrag}. To address this challenge, Graph-based RAG (Graph RAG) organizes facts through pre-built graph structures, enabling effective evidence traversal and multi-hop retrieval across interconnected knowledge units \cite{edge2024local}. However, such graphs often suffer from inherent limitations: they may be incomplete due to the difficulty of capturing all relevant relations, are costly to construct and maintain at scale, and become less effective when the underlying knowledge is frequently updated or evolving \cite{li2025subqrag}.

Recent advances such as LogicRAG \cite{chen2025you} reduce the reliance on pre-constructed graphs by decomposing complex queries into sub-questions and iteratively compressing retrieved evidence into a document-level memory, enabling multi-hop reasoning without explicit graph structures. However, such compression cannot fully filter out redundant or erroneous content, allowing noise to accumulate across iterations and gradually mislead subsequent reasoning \cite{zhuang-etal-2024-efficientrag}. Moreover, their query decomposition is often rigid and lacks adaptive control over granularity, so that simple questions tend to be over-decomposed while genuinely complex ones are not decomposed deeply enough, ultimately degrading reasoning performance \cite{kim-etal-2025-unirag}.

To overcome these limitations, we propose D$^2$F-ReAG, a multi-hop reasoning framework that performs on-demand question decomposition driven by reasoning reliability and augments generation with verified sub-question reasoning paths. Concretely, D$^2$F-ReAG continuously assesses the model's confidence in its current reasoning, and triggers further decomposition only when the reasoning over the current question is judged unreliable. Once a sub-question is solved with sufficient confidence and deemed relevant to the query, its reasoning trace is propagated upward to update and guide the reasoning of the root question, so that accurate intermediate results are effectively integrated rather than discarded or compressed away. To prevent over-reasoning and unnecessary computational overhead, D$^2$F-ReAG also adopts an early-stopping strategy that terminates the process as soon as the root question can be answered reliably. In summary, our contributions are:
\begin{itemize}
\item We propose D$^2$F-ReAG, a novel framework that dynamically decomposes complex questions into a sequence of logical sub-questions and progressively augments the reasoning process of the root question with the correct reasoning paths derived from these sub-questions, ensuring that accurate intermediate results are effectively propagated back to guide the final answer rather than being lost in lossy memory compression.
\item We dynamically and adaptively balance the decomposition depth according to question complexity through a confidence-driven control mechanism, enabling the framework to perform deeper decomposition only when necessary while avoiding redundant decomposition for simpler questions, thus achieving truly on-demand decomposition that adapts to varying reasoning difficulty.
\item We leverage reliable reasoning chains derived from sub-questions to iteratively update and correct the root reasoning process, which effectively identifies and fixes intermediate errors, thereby reducing error accumulation across reasoning steps and improving the overall correctness of the final answer.
\end{itemize}

\section{Related Work}


Graph-based methods organize external knowledge into structured graph representations to better support multi-hop retrieval and cross-document reasoning. GraphRAG~\cite{edge2024local} performs hierarchical community search by jointly leveraging local and global queries, allowing retrieval to operate at varying levels of granularity. LightRAG~\cite{guo-etal-2025-lightrag} further improves large-scale retrieval efficiency through a two-stage, graph-augmented indexing pipeline that balances coverage and cost. RAPTOR~\cite{sarthi2024raptor} constructs hierarchical summaries over the corpus to enable multi-granular retrieval at different levels of abstraction, thereby accommodating queries of diverse specificity. HippoRAG~\cite{gutiérrez2024hipporag} and its extension HippoRAG~2~\cite{gutiérrez2025ragmemorynonparametriccontinual} additionally enhance long-context coherence by ranking nodes with PageRank~\cite{page1999pagerank} and integrating paragraph-level memory, which helps preserve contextual continuity across retrieval steps. Despite these advances, graph-based methods generally rely on pre-constructed structures that are costly to build and maintain, and may become outdated when the underlying knowledge evolves.

Prompt-based methods, in contrast, interleave reasoning and retrieval through carefully designed prompts without modifying model parameters, offering a more flexible and lightweight alternative. ReAct~\cite{yao2023react} alternates between reasoning and retrieval in a step-by-step manner, enabling the model to dynamically decide when to retrieve based on its current reasoning state and observed evidence. ChainRAG~\cite{zhu-etal-2025-mitigating} decomposes complex questions into sub-questions guided by prompts, and subsequently performs iterative sentence-graph retrieval to capture both local fine-grained details and global document-level structures. LogicRAG~\cite{chen2025you} similarly decomposes questions into logical components and aggregates evidence across multiple reasoning steps, leveraging logical relationships among sub-questions to enhance retrieval coherence and reduce redundancy. SiGIR~\cite{chu-etal-2025-self} refines the reasoning process iteratively to progressively approach the final answer; however, it decomposes questions purely by atomicity and lacks explicit correctness verification during retrieval, which may allow erroneous intermediate results to propagate and accumulate across reasoning steps, ultimately degrading the quality of the final answer.

\section{Method}

\begin{figure*}[t]
  \centering
  \includegraphics[width=1.04\textwidth,trim=0.95cm 0.6cm 0.8cm 0.6cm,clip]{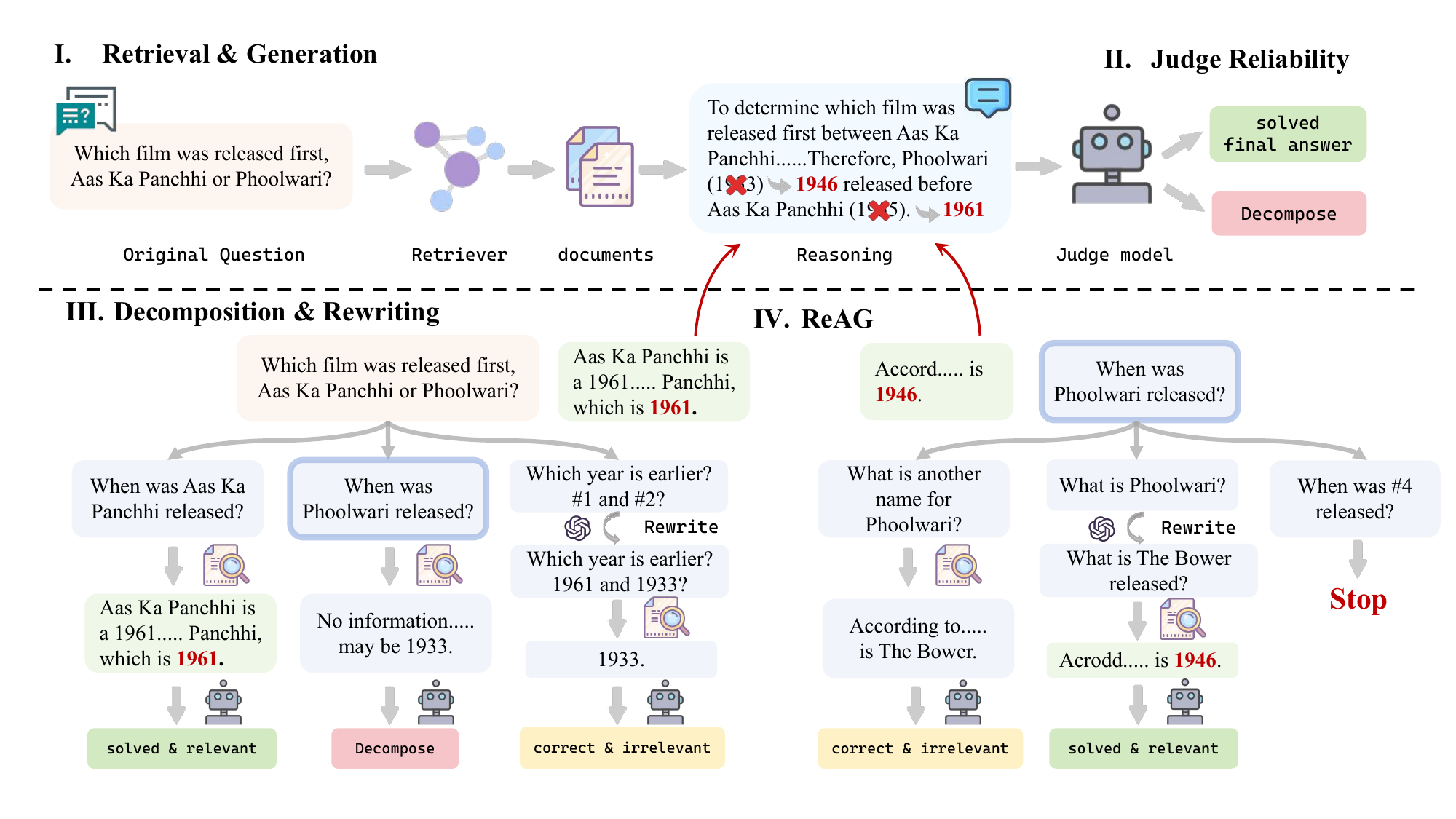}
    \caption{The D$^2$F-ReAG framework consists of four stages: (I) Retrieval \& Generation, which retrieves relevant documents to generate a reasoning. (II) Judge Reliability, where an LLM evaluates reasoning reliability against a threshold. (III) Decomposition \& Rewriting, which decomposes and rewrites the question when reliability is insufficient. (IV) ReAG (Reasoning-Augments generation), which integrates reliable and relevant sub-question reasoning to update the root reasoning. The process stops once the root question is answered reliably.}
  \label{fig:pdf_image}
\end{figure*}

We propose D$^2$F-ReAG, a dynamic framework for multi-hop reasoning. As shown in Figure~\ref{fig:pdf_image}, D$^2$F-ReAG begins with (I) Retrieval \& Generation, where first retrieves top-$k$ documents to generate question's reasoning. In (II) Judge Reliability, a judge model evaluates the reliability of the reasoning. If reliable, the question is solved; otherwise, (III) Decomposition \& Rewriting decomposes the question into sub-questions and rewrites them to retrieve more relevant evidence. Finally, (IV) ReAG (Reasoning-Augmented Generation) augments root reasoning with reliable, relevant reasoning traces of sub-questions. D$^2$F-ReAG stops once the root reasoning score exceeds a threshold.

\subsection{Retrieval \& Generation}

For the root question $q_{root}$ and each sub-question $q_i$, we retrieve the top-$k$ most relevant documents $D(q)$ from the corpus $\mathcal{C}$ using a dense retriever, and then generate the corresponding reasoning process $r(q)$ conditioned on the retrieved evidence. The retrieved documents serve as external knowledge that grounds the model's reasoning, mitigating hallucinations and providing factual support for downstream judgment. Formally, for any question $q \in \{q_{root}, q_1, \dots, q_n\}$,
\begin{equation}
r(q) = \mathrm{Generator}\bigl(q, D(q)\bigr),
\end{equation}
where $\mathrm{Generator}$ denotes the model that generates detailed reasoning based on relevant documents. By explicitly conditioning on $D(q)$, the generator is encouraged to perform evidence-aware reasoning rather than relying solely on its parametric memory, which is particularly important for knowledge-intensive multi-hop questions.

\subsection{Judge Reliability}
After generating the reasoning for the root or sub-questions, we judge its reliability using an LLM-based scoring mechanism. Rather than treating every generated reasoning chain as trustworthy, we introduce an explicit verification step that assesses whether the reasoning is logically coherent, factually consistent with the retrieved documents, and sufficient to answer the question. We denote the reliability score of the reasoning process as \( s_r(q) \), which is calculated by LLM:
\begin{equation}
    s_r(q) = \text{Score}(q, r(q)),
\end{equation}
where \( r(q) \) represents the reasoning process, and \( \text{Score}(q, r(q)) \) is the reasoning score provided by the LLM according to our predefined rubric. The rubric jointly considers multiple dimensions, including evidence grounding, logical consistency, and answer completeness, so that the resulting score reflects a holistic assessment of reasoning quality.

The reasoning score $s_r(q) \in [0,10]$ measures reasoning reliability (higher is better). We compare it to a threshold $\theta$ (set to 7 in our experiments): if $s_r(q) > \theta$, we consider the question solved. Otherwise, D$^2$F-ReAG thinks the reasoning unreliable and further decomposes the question for continued reasoning (Section~\ref{sec:decomposition}). This adaptive mechanism allows the system to allocate more computational effort only to questions that genuinely require deeper exploration, while terminating early on questions that are already well-addressed.
\begin{equation}
    \mathrm{Judge} =
    \begin{cases}
    \text{Solved}, & \text{if } s_r(q) > \theta \\
    \text{Decompose}, & \text{if } s_r(q) \leq \theta.
    \end{cases}
\end{equation}
If the current question is the root question and is deemed solved, the LLM generates the final answer based on the corresponding reasoning process, ensuring that the answer is directly traceable to verified reasoning steps:
\begin{equation}
a_{\text{root}} = \text{Answer}\!\left(r_{q_{\text{root}}}\right).
\label{eq:root_answer}
\end{equation}
If a sub-question is reliably solved, we use its relevant sub-questions to update the root reasoning in ReAG (Section~\ref{sec:reag}), so that the verified intermediate conclusions can progressively enrich the global reasoning context.

\subsection{Decomposition \& Rewriting}
\label{sec:decomposition}
When the root question or any sub-question remains unsolved, we use prompt engineering to logically decompose it into smaller, more tractable sub-questions, each focusing on a single reasoning hop or a narrower information need. This divide-and-conquer strategy reduces the cognitive load on the generator and makes retrieval more targeted.
\begin{equation}
sub(q) = \text{Decompose}(q),
\end{equation}
where $sub(q)$ is the set of sub-questions obtained by logically decomposing the question $q$.

D$^2$F-ReAG first solves each sub-question in order and judges the reliability of the reasoning. If reliable, we rewrite the related other sub-questions $q \in \mathcal{S}_i$ based on the correct reasoning $r_q$, so that subsequent sub-questions can leverage previously verified intermediate results and avoid redundant or contradictory reasoning paths.
\begin{equation}
\mathcal{S}_i' = \{\, \text{Rewrite}(q, r_q) \mid q \in \mathcal{S}_i \,\},
\label{eq:rewrite}
\end{equation}
where $r_q$ is the reasoning to the current question, and $q_i$ is the related sub-question. Through this rewriting step, ambiguous references and missing entities in the original sub-questions can be resolved using the newly acquired evidence, leading to higher-quality retrieval in the next iteration.

\subsection{ReAG}
\label{sec:reag}
When a sub-question is solved reliably, we check its relevance to the root question and use its reasoning to update the root reasoning if relevant. This relevance check prevents tangential or off-topic sub-question results from contaminating the global reasoning state.
\begin{equation}
\text{Check}(q_i, q) = 
    \begin{cases}
    \text{relevant} \\
    \text{irrelevant} 
    \end{cases}
\end{equation}
where Check$(q_i, q)$ denotes the relevance of $q_i$ to the root question $q$.

If a sub-question's reasoning is reliable and relevant, it is used to update the root reasoning. The update operation integrates newly verified evidence and intermediate conclusions into the existing root reasoning, gradually building a more complete and coherent chain of thought toward the final answer. Formally, this can be written as:
\begin{equation}
r'(q_{root}) = \text{Update}(r(q_{root}), r(q)),
\label{eq:update}
\end{equation}
where \( r'(q_{root}) \) is the updated reasoning process of the root problem \( q \), \( r(q) \) is the reasoning process of the sub-question \( q_i \), and \( \text{Update}(r(q_{root}), r(q)) \) updates \( r(q_{root}) \) by incorporating \( r(q) \). 

Once the root reasoning is reliable, we stop processing the remaining sub-questions and obtain the final answer to avoid over-reasoning. This early-stopping strategy not only reduces unnecessary computational overhead but also prevents the introduction of noise from over-decomposition, which could otherwise dilute the focus of the root reasoning and degrade answer accuracy.

\begin{table*}[t]
  \centering
  \caption{Performance comparison on HotpotQA, 2WikiMultiHopQA, and MuSiQue using Str-Acc and LLM-Acc. $\dagger$ denotes our reimplementation; the others are from LogicRAG. Best results are in bold, and second-best results are underlined.}
  \footnotesize
  \renewcommand{\arraystretch}{0.8}
  \setlength{\tabcolsep}{2.8pt}
  \begin{tabular}{l l c c c c c c}
    \toprule
    \multirow{2}{*}{\textbf{Type}} & \multirow{2}{*}{\textbf{Method}} &
    \multicolumn{2}{c}{\textbf{HotpotQA}} &
    \multicolumn{2}{c}{\textbf{2Wiki}} &
    \multicolumn{2}{c}{\textbf{MuSiQue}} \\
    \cmidrule(lr){3-4} \cmidrule(lr){5-6} \cmidrule(lr){7-8}
    & & \textbf{Str.} & \textbf{LLM} & \textbf{Str.} & \textbf{LLM} & \textbf{Str.} & \textbf{LLM} \\
    \midrule

    \multirow{4}{*}{Zero-shot}
    & Llama3 (8B)       & 17.1 & 11.1 & 22.3 & 4.7  & 2.3  & 2.0  \\
    & Llama3 (13B)      & 23.7 & 20.1 & 33.8 & 15.4 & 6.4  & 6.0  \\
    & GPT-3.5-Turbo     & 31.5 & 35.4 & 24.0 & 22.0 & 7.9  & 10.9 \\
    & GPT-4o-Mini       & 38.7 & 36.3 & 26.4 & 24.3 & 17.6 & 14.0 \\
    \midrule

    \multirow{5}{*}{Graph}
    & RAPTOR            & 48.1 & 57.8 & 47.7 & 45.9 & 25.2 & 29.1 \\
    & GraphRAG          & 39.6 & 45.2 & 46.3 & 43.3 & 16.5 & 19.4 \\
    & LightRAG          & 47.8 & 57.7 & 43.1 & 36.3 & 18.1 & 19.4 \\
    & HippoRAG          & 53.5 & 56.6 & 47.2 & 47.2 & 24.9 & 30.1 \\
    & HippoRAG2         & \textbf{56.7} & 61.9 & 50.0 & 47.1 & 27.0 & 32.6 \\
    \midrule

    \multirow{4}{*}{Prompt-RAG}
    & VanillaRAG        & 43.2 & 53.1 & 43.0 & 42.0 & 20.3 & 23.6 \\
    & ReAct$^\dagger$   & 54.2 & 56.5 & 54.9 & 50.8 & 28.8 & 31.9 \\
    & ChainRAG$^\dagger$& 52.1 & 56.6 & \underline{68.9} & \underline{66.2} & \underline{31.0} & 33.8 \\
    & LogicRAG$^\dagger$& 54.2 & \underline{62.5} & 65.3 & 62.6 & 29.6 & \underline{36.5} \\
    \midrule

    Ours & D$^2$F-ReAG  & \underline{55.8} & \textbf{63.4} & \textbf{70.3} & \textbf{68.9} & \textbf{32.2} & \textbf{37.9} \\
    \bottomrule
  \end{tabular}

  \label{tab:accents}
\end{table*}

\section{Experiments}
\subsection{Dataset and Metrics}
We evaluate on three standard multi-hop reasoning benchmarks: HotpotQA \cite{yang-etal-2018-hotpotqa}, MuSiQue \cite{trivedi2021musique} and 2WikiMultiHopQA \cite{ho-etal-2020-constructing}, covering diverse cross-document and multi-hop reasoning. Following HippoRAG~2 \cite{gutiérrez2025ragmemorynonparametriccontinual}, we use the same retrieval corpus and randomly sample 1,000 questions from each validation set for evaluation, ensuring fair comparison.

Following LogicRAG~\cite{chen2025you}, we adopt two metrics: \textbf{Str-Acc} and \textbf{LLM-Acc}. \textbf{Str-Acc} measures lexical correctness by checking whether the prediction exactly matches the ground-truth answer after standard normalization. \textbf{LLM-Acc} instead employs a strong LLM as an automatic judge to assess whether the prediction is semantically equivalent to the reference, tolerating paraphrasing and surface-form variations. Reporting both metrics enables a balanced evaluation of strict lexical matching and flexible semantic correctness.

\subsection{Baselines}
We compare D$^2$F-ReAG with three categories of baselines that cover representative paradigms in multi-hop question answering. \textbf{Zero-shot} baselines directly prompt LLMs without any external retrieval, including LLaMA3 (8B) and LLaMA3 (13B) \cite{dubey2024llama}, as well as gpt-3.5-turbo and gpt-4o-mini \cite{achiam2023gpt}, which serve to reflect the intrinsic reasoning ability of LLMs. \textbf{Graph RAG} baselines incorporate pre-constructed graph structures to support multi-hop retrieval, including RAPTOR \cite{sarthi2024raptor}, GraphRAG \cite{edge2024local}, LightRAG \cite{guo-etal-2025-lightrag}, HippoRAG \cite{gutiérrez2024hipporag}, and HippoRAG~2 \cite{gutiérrez2025ragmemorynonparametriccontinual}. \textbf{Prompt-based RAG} baselines instead interleave reasoning and retrieval through prompt design without relying on explicit graph construction, including ReAct \cite{yao2023react}, ChainRAG \cite{zhu-etal-2025-mitigating}, and LogicRAG \cite{chen2025you}.

\subsection{Implementation Details.} For fair comparison, all methods adopt the same experimental configuration. We use sentence-transformers/all-MiniLM-L6-v2~\cite{wang-etal-2021-minilmv2} as the unified embedding model for dense retrieval. The number of retrieved passages (top-$k$) is fixed to 3 across all methods. For answer generation, we employ gpt-4o-mini~\cite{achiam2023gpt} as the backbone LLM, ensuring that performance differences stem from the retrieval and reasoning strategies rather than the underlying generator. All experiments are conducted on a single NVIDIA RTX 3090 GPU.

\subsection{Main Results}


Table~\ref{tab:accents} reports results on three multi-hop reasoning benchmarks (HotpotQA, 2WikiMultiHopQA, and MuSiQue). D$^2$F-ReAG achieves the best or near-best performance on both Str-Acc and LLM-Acc across all three datasets, demonstrating consistent superiority over diverse baselines.

Zero-shot LLMs benefit from stronger backbones, with gpt-4o-mini clearly surpassing LLaMA3 (8B/13B), yet even the best zero-shot model trails retrieval-based methods, underscoring the need for external evidence in multi-hop QA. Graph-based RAG methods (RAPTOR, GraphRAG, LightRAG, HippoRAG, HippoRAG~2) improve performance by structuring context into explicit graphs, with HippoRAG~2 being the strongest (e.g., 56.7 Str-Acc on HotpotQA), though their gains shrink on harder datasets like MuSiQue where graphs miss multi-hop relations. Prompt-based RAG methods interleave retrieval and reasoning: ChainRAG performs well on 2Wiki via sub-question decomposition with sentence-graph retrieval, and LogicRAG attains the best baseline LLM-Acc on HotpotQA (62.5) through logical decomposition, but both suffer from fixed decomposition and noise accumulated over iterations.

D$^2$F-ReAG consistently outperforms all baselines on 2Wiki (70.3 / 68.9) and MuSiQue (32.2 / 37.9), and achieves the highest LLM-Acc on HotpotQA (63.4), with the largest gains on 2WikiMultiHopQA where deep multi-hop reasoning is most needed. Against the strongest prompt-based baseline LogicRAG, it yields up to 5.4-point Str-Acc and 6.4-point LLM-Acc improvements on 2Wiki, showing that on-demand decomposition and reliability-guided reasoning mitigate the over- or under-decomposition of fixed-depth methods. Joint gains on both metrics further indicate that the recovered reasoning paths are not only lexically aligned with gold answers but also semantically more faithful. We report the best scores across runs, and provide a case study in Appendix~\ref{sec:appendix} to illustrate how D$^2$F-ReAG corrects intermediate errors via reliable sub-question reasoning.
\begin{figure}[t]
\centering
\begin{tikzpicture}
\begin{axis}[
    ybar,
    bar shift auto,
    width=0.72\columnwidth,
    height=0.55\columnwidth,
    ymin=0, ymax=1,
    ytick={0,0.2,0.4,0.6,0.8},
    ylabel={rate},
    ymajorgrids,
    grid style={dashed,gray!45},
    xticklabel style={font=\footnotesize},
    axis lines=box,
    tick align=inside,
    tick style={black, line width=0.6pt},
    major tick length=3pt,
    symbolic x coords={MuSiQue,2Wiki,HotpotQA},
    xtick=data,
    enlarge x limits=0.18,
    bar width=7pt,
    legend style={
        at={(0.02,0.98)},
        anchor=north west,
        legend columns=1,
        draw=black,
        fill=white,
        font=\scriptsize,
        inner xsep=2pt,
        inner ysep=1pt
    },
    nodes near coords={\pgfmathprintnumber[fixed,precision=2]{\pgfplotspointmeta}},
    point meta=y,
    every node near coord/.append style={font=\scriptsize, yshift=1pt,text=black},
]

\addplot+[fill=teal!25, draw=teal!50!black] coordinates {
    (MuSiQue,0.59)
    (2Wiki,0.632)
    (HotpotQA,0.799)
};
\addlegendentry{w/o decomposition}

\addplot+[fill=purple!20, draw=purple!70!black] coordinates {
    (MuSiQue,0.41)
    (2Wiki,0.368)
    (HotpotQA,0.201)
};
\addlegendentry{w/\ decomposition}

\end{axis}
\end{tikzpicture}
\caption{Solved with vs. without decomposition on three benchmarks.}
\label{fig:pdf_image_rate}
\end{figure}
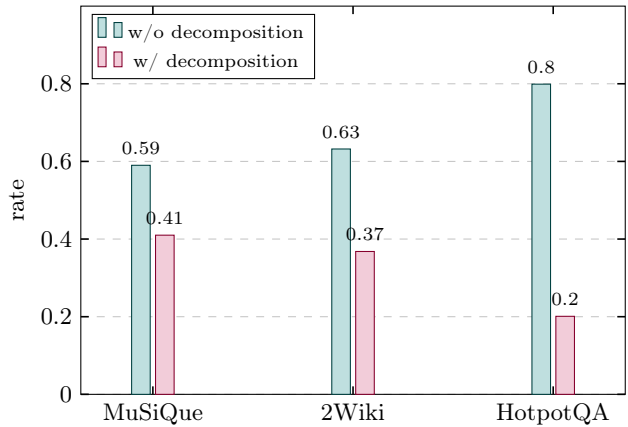

\begin{table}[htbp]
\centering
\caption{Accuracy and efficiency comparison on 2WikiMultiHopQA (with vs. without decomposition).}
\renewcommand{\arraystretch}{0.9}
\setlength{\tabcolsep}{3.5pt}
\begin{tabular}{l c c c c}
\toprule
\textbf{Method}
& \makecell{\textbf{Str}\\\textbf{Acc}}
& \makecell{\textbf{LLM}\\\textbf{Acc}}
& \makecell{\textbf{Avg.}\\\textbf{Time (s)}}
& \makecell{\textbf{Avg.}\\\textbf{Tokens}} \\
\midrule
ReAct        & 54.9 & 50.8 & 13.93 & 11287 \\
LogicRAG     & 65.3 & 62.6 & 15.35 & 1998  \\
\midrule
Ours (No D)  & 71.9 & 71.0 & 9.49  & 2267  \\
Ours (+D)    & 67.3 & 65.0 & 73.54 & 13321 \\
Ours (Avg.)  & 70.3 & 68.9 & 32.4  & 6057  \\
\bottomrule
\end{tabular}

\label{tab:2wiki_efficiency}
\end{table}
\subsection{Decomposition \& Efficiency Comparison}
Figure \ref{fig:pdf_image_rate} shows that a large proportion of questions can be solved without decomposition, demonstrating the necessity of dynamic decomposition to efficiently handle both simple and complex questions.\\
Since ChainRAG also uses a pre-built sentence graph, we compare latency and efficiency with ReAct and LogicRAG. Table \ref{tab:2wiki_efficiency} shows that always decomposing and judging reliability greatly increases time and token cost. In contrast, on-demand decomposition skips unnecessary steps, speeding up easy cases while still decomposing for hard ones. While our multi-step judgments can take longer on some difficult examples, Figure \ref{fig:pdf_image_rate} confirms these long cases are relatively few. Overall, D$^2$F-ReAG balances between performance and efficiency and better matches human-like reasoning.

\section{Conclusion}
Existing RAG methods still struggle with multi-hop questions, primarily due to their fixed decomposition strategies and limited ability to filter out erroneous or irrelevant information during reasoning. To address these issues, we propose D$^2$F-ReAG (Dynamic Decomposition and Filtering for Multi-Hop Reasoning-Augmented Generation), a framework that adaptively decomposes questions only when necessary and selectively filters intermediate reasoning according to its reliability. By leveraging verified reasoning from reliable sub-questions to progressively refine the root-level generation, D$^2$F-ReAG effectively suppresses error propagation and produces more faithful reasoning chains, achieving strong performance across challenging multi-hop benchmarks.

\section{Case Study}
\label{sec:appendix}

We present a case study comparing the reasoning behaviors of \textbf{D$^2$F-ReAG} and \textbf{LogicRAG}.

\textbf{Question:} Which film has the director who died later, \textit{45 Calibre Echo} or \textit{Bons Baisers De Hong Kong}?\\
\textbf{Gold Answer:} \textit{Bons Baisers De Hong Kong}\\

\textbf{Failure Case of LogicRAG.}\\
\textbf{Model Answer:} \textcolor{red}{\textit{45 Calibre Echo}} (\textbf{Incorrect})\quad\textbf{Rounds:} 1

\paragraph{Retrieved Contexts.}
LogicRAG retrieves three biography-style passages:
\begin{itemize}
    \item Bruce M. Mitchell: includes his death date (September 26, 1952), supporting reasoning about \textit{45 Calibre Echo}.
    \item John Edward Bruce / Fred Bradley (rower): unrelated to either queried film.
\end{itemize}

\paragraph{Dependency Analysis (sorted).} Death date of Bruce M.~Mitchell; Death date of Yvan Chiffre.

\paragraph{Round 1.}
\textbf{Query:} Death date of Bruce M.~Mitchell.\quad\textbf{can\_answer:} true.\\
\textbf{Understanding:} Bruce M.~Mitchell (director of \textit{45 Calibre Echo}) died on September 26, 1952; the death date of Yvan Chiffre is not retrieved. \textcolor{red}{LogicRAG nevertheless concludes \textit{45 Calibre Echo} due to the missing comparison target.}

\vspace{6pt}
\textbf{Correct Case of D$^2$F-ReAG.}\\
\textbf{Model Answer:} \textcolor{green}{\textit{Bons Baisers De Hong Kong}} (\textbf{Correct})\\
\textbf{Root Rationale:} The retrieved context names the directors (Bruce M.~Mitchell and Yvan Chiffre) but does not provide their death dates. \textbf{Judging reliability:} 0.4

\textbf{Decompose \& Rewriting:}
\begin{itemize}
    \item Who directed \textit{45 Calibre Echo} and when did the director die?
    \item Who directed \textit{Bons Baisers De Hong Kong} and when did the director die?
    \item Which year is later?
\end{itemize}

\textbf{Filtering for Multi-Hop Reasoning-Augmented Generation.}\\
\textbf{Subq 1:} Director of \textit{45 Calibre Echo} is Bruce M.~Mitchell. \textbf{Reliability:} 0.8\\
\textbf{Subq 2:} Director of \textit{Bons Baisers De Hong Kong} is Yvan Chiffre. \textbf{Reliability:} 0.8

\textbf{Updating Root Reasoning and Rewriting.}\\
Rewritten sub-questions: When did Bruce M.~Mitchell die? When did Yvan Chiffre die?

\textbf{Filtering for Multi-Hop Reasoning-Augmented Generation.}\\
Yvan Chiffre died on 1990-01-01; Bruce M.~Mitchell died on 1988-02-19.

\textbf{Updating Root Reasoning and Rewriting.}\\
\textbf{Rewritten sub-question 3:} Which year is later, 1988 or 1990?\\
\textbf{Subq 3:} 1990. \textbf{Reliability:} 1.0\\
\textbf{Final Answer:} \textcolor{green}{\textit{Bons Baisers De Hong Kong}}

\bibliographystyle{splncs04}
\bibliography{references}
\end{document}